\documentclass[10pt,twocolumn,letterpaper]{article}

\usepackage{iccv}
\usepackage{times}
\usepackage{epsfig}
\usepackage{graphicx}
\usepackage{amsmath}
\usepackage{amssymb}

\usepackage{makecell}
\usepackage{multirow}
\usepackage{diagbox}

\usepackage[pagebackref=true,breaklinks=true,letterpaper=true,colorlinks,bookmarks=false]{hyperref}

\iccvfinalcopy % *** Uncomment this line for the final submission

\def\iccvPaperID{6752} % *** Enter the ICCV Paper ID here

\ificcvfinal\fi

\begin{document}

%%%%%%%%% TITLE
\title{Linguistic Query-Guided Mask Generation for Referring Image Segmentation}

% \and
\author{Zhichao Wei$^{1}$\hspace{1cm}Xiaohao Chen$^{1}$\hspace{1cm}Mingqiang Chen$^{1}$\hspace{1cm}Siyu Zhu$^{1*}$\\
${}^{1}$Alibaba Group\
}
\maketitle
\def\thefootnote{*}\footnotetext{Siyu Zhu is the corresponding author.}

\ificcvfinal\thispagestyle{empty}\fi

%%%%%%%%% ABSTRACT
\begin{abstract}
   Referring image segmentation aims to segment the image region of interest according to the given language expression, which is a typical multi-modal task. 
   Existing methods either adopt the pixel classification-based or the learnable query-based framework for mask generation, both of which are insufficient to deal with various text-image pairs with a fix number of parametric prototypes. In this work, we propose an end-to-end framework built on transformer to perform Linguistic query-Guided mask generation, dubbed LGFormer. It views the linguistic features as query to generate a specialized prototype for arbitrary input image-text pair, thus generating more consistent segmentation results. Moreover, we design several cross-modal interaction modules (\eg, vision-language bidirectional attention module, VLBA) in both encoder and decoder to achieve better cross-modal alignment.
%   The proposed framework is simple and effective. 
   Extensive experiments demonstrate that our LGFormer achieves a new state-of-the-art performance on ReferIt, RefCOCO+, and RefCOCOg by large margins.

%   The linguistic features are viewed as the query to generate a prototype, grouping pixels with high responses into segmentation mask directly. Concretely, we use linguistic features to generate a language-attended query as the input of the transformer decoder and use multi-scale visual features to update it sequentially. 
%   In this manner, the query is able to capture crucial information of the unique referred object in the image, and play the role of referent center embedding to generate the segmentation mask from mask features directly. In addition, different from the existing encoder-fusion or decoder-fusion methods, we perform cross-modal fusion in both stages by the well-designed modules to align semantic representations of vision and language modalities. 
\end{abstract}

%%%%%%%%% BODY TEXT
\section{Introduction}

Referring image segmentation (RIS) aims to generate a segmentation mask for the target object corresponding to a given language expression. In contrast to traditional segmentation tasks~\cite{fu2019dual, he2019adaptive, he2017mask, long2015fully} which can only process a predefined set of categories, referring image segmentation is no longer limited to specific classes and has wider applications including human-robot interaction~\cite{wang2019reinforced} and interactive photo editing~\cite{chen2018language}. 
{\begin{figure}[!htb]
\centering
\includegraphics[scale=0.45]{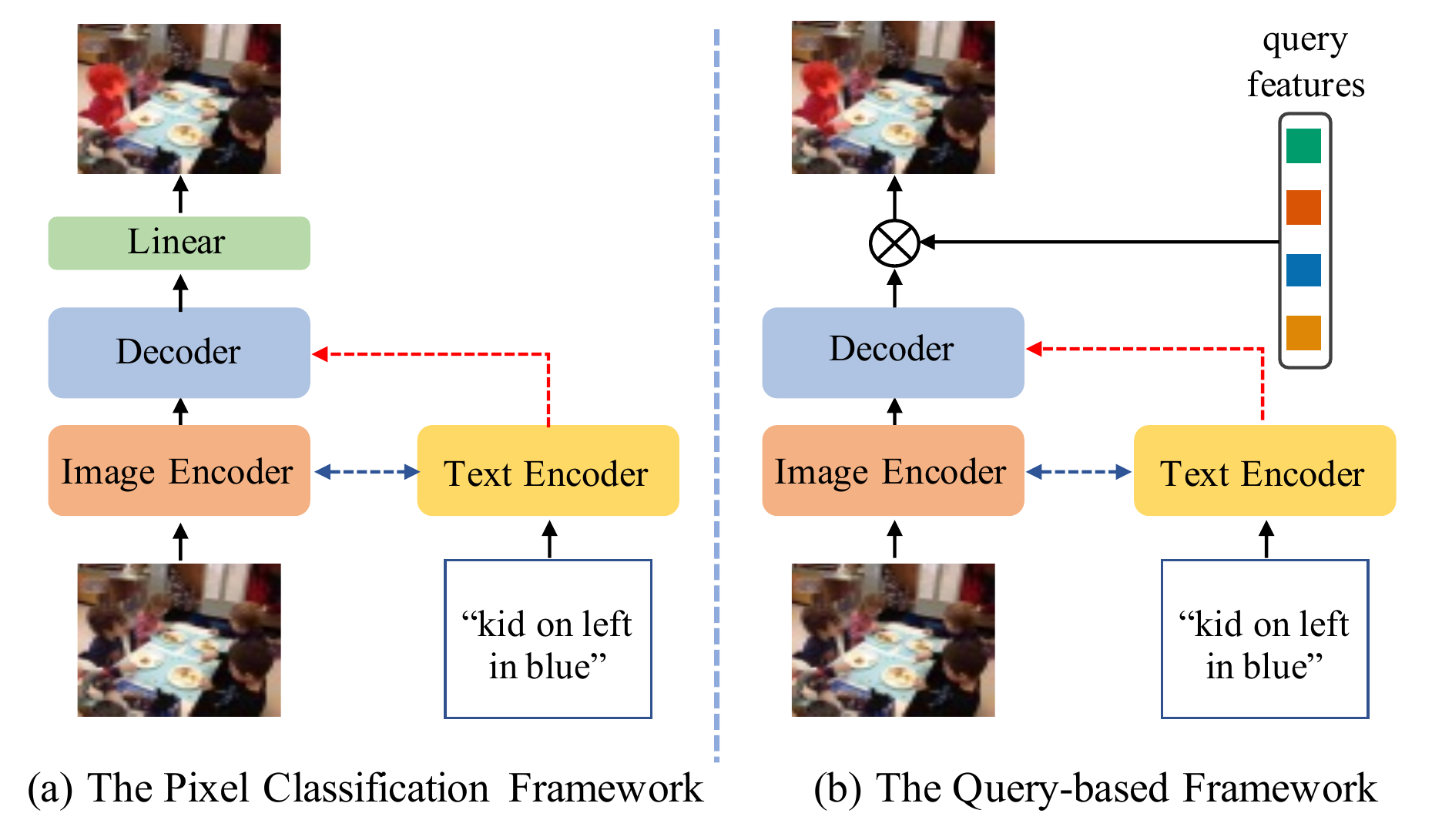}
\caption{Illustration of the main architectures for referring image segmentation. 
(a) Pixel classification framework. 
(b) Query-based framework. Both \cite{zhang2022coupalign} and our LGFormer belongs to the latter. }
\vspace{-0.5cm}
\label{Fig.design_principles} %用于文内引用的标签
\end{figure}}

Although recent methods have achieved remarkable performance in this area, referring image segmentation is still a challenging task. The first challenge is \textit{the alignment of cross-modal features}. As vision and language modality naturally maintain different attributes, aligning pixel-level visual features to corresponding holistic linguistic features is difficult. Another challenge is \textit{the variance of text-image pairs}. The same language expression associated with different images could yield completely different segmentation results, while different language expressions related to the same image could produce the same segmentation results. Each text-image pair belongs to a unique class, making RIS an open-class segmentation task.

% Following the design principles of traditional semantic segmentation works, 
Most of the existing RIS methods are based on \textit{pixel classification} framework, which uses a linear classifier to generate segmentation masks, as shown in Fig.~\ref{Fig.design_principles}(a). 
% To improve the segmentation accuracy, these methods investigate various fusion strategies for better cross-modal feature alignment. 
Previous methods~\cite{chen2019see, ding2021vision, hu2020bi, huang2020referring, hui2020linguistic, ye2019cross, li2018referring, liu2017recurrent, margffoy2018dynamic, wu2022towards, wang2022cris, liu2021cross} often employ the decoder-fusion strategy, which first extracts visual and linguistic features from two independent uni-modal encoders respectively, then fuses these multi-modal representations into the same embedding space.
% and finally generates the target mask. 
% However, these methods only interact with high-level features of two modalities and miss the importance of low-level visual features such as texture, shape, and color, as well as word features. 
In contrast, some recent works~\cite{feng2021encoder, li2021mail, yang2022lavt} adopt the encoder-fusion strategy,
% to solve the referring image segmentation problem, 
which performs cross-modal interactions in the early encoder stage.
% to update visual features with linguistic features. 
Although \textit{pixel classification}-based methods have demonstrated their effectiveness, they mainly focus on resolving the issue of cross-modal features alignment but disregard the challenges that arise from the variance of text-image pairs in this task. From a prototype view, the pixel classification framework views the weights of the linear classifier as the prototype which groups pixels with high responses into segments~\cite{zhou2022rethinking}. However, one single fixed prototype is insufficient to describe the rich variance of text-image pairs. As shown in the first column of Fig. \ref{Fig.text_to_images}, these methods often fail to exploit the details (such as positions and attributes) and produce over- or under-segmented results with longer language expressions.
% they still have some intrinsic limitations. First, they failed to capture instance-level information and cannot perceive the discrepancy of different inputs since the parameters of the linear classifier will be fixed after training. Second, both decoder-fusion and encoder-fusion strategies are applied in the form of aggregating linguistic features into visual features to obtain text-aligned visual features but neglected the role of aligning linguistic features with visual features. 
% We argue that this scheme is insufficient to address the variety of inputs images and languages.

{\begin{figure}[!htb]
\centering
\includegraphics[scale=0.65]{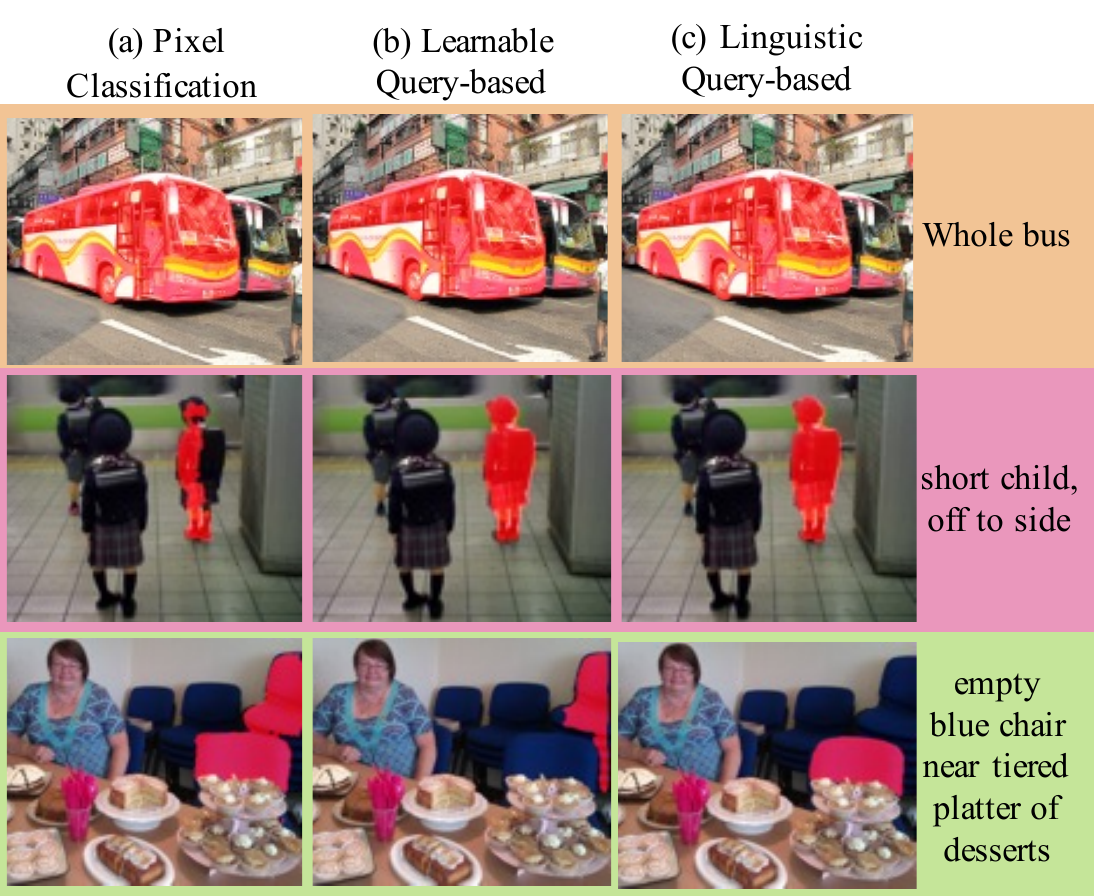}
\caption{(a) The pixel classification-based methods perform well in easy cases but often produce inaccurate results with long language expressions. (b) The learnable query-based methods can capture more detailed information but still fail in extremely complicated language expression cases. (c) The linguistic query-based methods still generate high-quality segmentation results for both easy and complicated cases.}
\label{Fig.text_to_images} %用于文内引用的标签
\vspace{-0.2cm}
\end{figure}} 

% These limitations motivate us to develop a simple yet effective framework to solve the RIS task. 
Recently, \textit{query-based} methods~\cite{cheng2022masked, cheng2021per, yu2022k} have greatly boosted the performance over the previous methods in the traditional image segmentation task.
These methods introduce a fixed number of learnable queries as the input of the transformer decoder and associate each query with an instance. Based on this framework, the very recent work~\cite{zhang2022coupalign} achieved promising results on the RIS task, whose architecture is shown in Fig.~\ref{Fig.design_principles}(b). 
% The recent success of recent query-based segmentation methods motivates us to develop a simple yet effective framework for RIS. 
% Since the instance-level information is captured by queries, 
However, we argue that applying the learnable query-based framework to the RIS task directly is sub-optimal. Specifically, although the prototypes can be updated online with the multi-modal features, they are initialized with a set of fixed queries that are irrelevant to the input text-image pair, which greatly hinders the alignment with pixel embeddings and may generate inconsistent segmentation results.
% Second, overmuch learnable queries for the unique referred object in the image are redundant, which makes the model confused to associate multiple queries to one object. 
As is shown in the second column of Fig. \ref{Fig.text_to_images}, the learnable query-based method can capture some details with online updated prototypes and generate the more accurate results, but are still insufficient to deal with the complicated language expressions.
% for the insufficient utilization of linguistic features which contain the unique clues to find the image region of interest.

% The recent breakthroughs in query-based segmentation methods motivate us to develop a straightforward but efficient framework for RIS.
To extend this paradigm to the RIS task elegantly, we propose the notion of linguistic query and introduce LGFormer, an end-to-end framework for referring image segmentation. It utilizes the linguistic features as the query to generate a specialized prototype for the input text-image pair, which will group pixels with high responses into mask directly.
% propose LGFormer, an end-to-end framework for referring image segmentation that views linguistic features as the prototype and generates a segmentation mask by grouping pixels with high responses. 
% In this manner, the linguistic-attended query is able to capture the instance-level visual information from the mask features and focus on the referred object, thus generating more consistent segmentation results. 
Different from previous query-based segmentation methods, one query is enough in our LGFormer for the attendance of linguistic information. Moreover, to obtain more discriminative feature representations, we design a vision-language bidirectional attention (VLBA) module and a cross-modal decoder, performing cross-modal interactions at both the encoding and decoding stages.

To summarize, our main contributions are as follows:
\begin{itemize}
    % \item We propose LGFormer, a transformer-based RIS framework that first introduces language-attended query for mask generation. Given an image and the corresponding language expression, the proposed LGFormer directly groups image pixels with high responses to only one linguistic query into a mask without any matching or integration mechanisms. 
    % \item We design the vision-language bidirectional attention module (VLBA) and cross-modal decoder for encoder-fusion and decoder-fusion respectively, which ensure sufficient cross-modal interactions for accurate mask generation.
    \item We introduce the linguistic query-guided mask generation mechanism to solve referring image segmentation for the first time. Given arbitrary text-image pair, we use the linguistic features as the query to generate a specialized prototype by interactions with visual features for mask generation.
    \item Based on the proposed mask generation mechanism, we design a simple yet effective framework, dubbed LGFormer. It use the specialized prototype to group image pixels with high responses into segmentation mask directly in an end-to-end manner. Moreover, we design several cross-modal interaction modules for better cross-modal alignment.
    \item The proposed method achieves state-of-the-art results on multiple datasets, including ReferIt\cite{kazemzadeh2014referitgame}, RefCOCO+~\cite{yu2016modeling}, and RefCOCOg~\cite{mao2016generation}, by large margins without bells and whistles.
\end{itemize}

\section{Related Work}

% {\begin{figure*}[!htb]
% \centering
% \includegraphics[scale=0.47]{figures/pipeline.pdf}
% \caption{An overview of the LGFormer framework. It consists of a vision-language encoder with our proposed VLBA module for cross-modal interactions, a vision-language decoder with a pixel decoder for integrating multi-scale visual features, and a transformer decoder for generating the visual-attended linguistic features as the referent center embedding.}
% \vspace{-0.3cm}
% \label{Fig.pipeline} %用于文内引用的标签
% \end{figure*}}

\paragraph{Pixel Classification Framework.} Motivated by the success of \textit{pixel classification} framework in traditional semantic segmentation, most of the existing methods adopt the FCN-like head to decode the mask. Early works mainly focused on fusing multi-modal representations on the decoder. Hu \etal~\cite{hu2016segmentation} proposed to utilize CNN and RNN to extract visual and linguistic features separately and then obtain a segmentation mask through the concatenation-convolution operation. Based on this pipeline, some works~\cite{liu2017recurrent, chen2019see, hu2020bi, ye2020dual} improved the performance by utilizing more powerful feature encoders and designing more ingenious fusion strategies (\eg, recurrent fusion). With the success of attention mechanisms in the communities of natural language processing and computer vision, follow-up works~\cite{ye2019cross, luo2020cascade, ding2021vision, kim2022restr} attempted to model intra-modal and cross-modal relationships by adopting self-attention and cross-attention operations. For example, Ye \etal ~\cite{ye2019cross} proposed Cross-Modal Self-Attention (CMSA) to highlight informative visual and linguistic elements. Ding \etal~\cite{ding2021vision} used vision-guided attention to generate multiple linguistic queries which understand language expression from different aspects. Kim \etal~\cite{kim2022restr} conducted both intra-modality and inter-modality interactions by self-attention operations, alleviating the weakness of CNN in capturing language information and cross-modal information. Recent studies~\cite{feng2021encoder, kamath2021mdetr, yang2022lavt} have shown that cross-modal interactions during feature extraction can further enhance multi-modal alignment. Feng \etal~\cite{feng2021encoder} replaced the vision encoder with a multi-modal encoder by adopting an early cross-modal interaction strategy, which achieved deep interweaving between visual and linguistic features. Yang \etal~\cite{yang2022lavt} adopted language-aware visual attention during feature encoding and effectively exploited the transformer encoder for modeling multi-modal context. Instead of using a uni-modal pre-trained backbone for features extraction, Wang \etal~\cite{wang2022cris} transfers multi-modal knowledge of the contrastive language-image pretraining (CLIP)~\cite{radford2021learning} for better cross-modal alignment.

\paragraph{Query-based Framework.} 
% With the long-range modeling ability of the attention mechanism, the transformer architecture has dominated most computer vision tasks, such as image classification~\cite{dosovitskiy2020image, wang2021pyramid, liu2021swin} and object detection~\cite{carion2020end, zhu2020deformable}.
Some recent image segmentation works~\cite{wang2021max, cheng2021per, cheng2022masked, yu2022k} employ transformers with additional cross-attention as mask decoder which generates segmentation masks by object queries (\textit{i.e.}, learnable embeddings). Wang \etal~\cite{wang2021max} proposed an end-to-end panoptic segmentation method named Max-DeepLab which employs a mask decoder to generate dynamic filters. Cheng \etal~\cite{cheng2021per} proposed MaskFormer which shows good performance in both semantic- and instance-level segmentation tasks in a unified manner. By using mask attention, multi-scale high-resolution features, and optimizing the training strategy, Mask2Former~\cite{cheng2022masked} outperformed specialized architectures across different segmentation tasks by a universal image segmentation architecture. From the perspective of clustering, Yu \etal~\cite{yu2022k} proposed kMax-DeepLab, endowing better interpretation ability of mask decoder. 
The very recent work CoupAlign~\cite{zhang2022coupalign} is also based on the query-based framework. However, this method introduces a set of (i.e., 100) learnable queries to obtain mask proposals and generate masks by additional integration operations. 
% which may make the model confused to associate multiple learnable queries to the unique referred object.
% Different from existing \textit{pixel classification}-based methods, we propose a linguistic \textit{query-based} method that achieves a remarkable performance gain with a simple yet efficient framework. 
% , thus generating inconsistent or even wrong segmentation results.
We argue that adopting the learnable query for the RIS task is sub-optimal, as the learnable query is irrelevant to the input.
% and multiple learnable queries could make the model confused for the unique referred object.
To extend this query-based paradigm into referring image segmentation task, we propose to treat the vision-attended linguistic features as the unique query for generating a linguistic prototype to group pixels into segments. 

\section{Methodology}
In this section, we first overview two frameworks used in existing methods from a prototype view, including the pixel classification framework and the query-based framework, and describe how we improve the latter to fit the referring image segmentation task. Then, we introduce the network structure and details of the proposed LGFormer. 

\subsection{Linguistic Prototype}

From a prototype view, the process of segmentation can be seen as utilizing the prototype $\rho_k$ to group pixel embeddings $E_i \in \mathbb{R}^{D}$ with high response and generate the mask\cite{zhou2022rethinking}, which can be formulated as follow:

\begin{equation}
\label{equ:prototype_view}
p \left ( k|E_i \right ) = \frac{\exp (\rho^{T}_k E_i)}{ {\textstyle \sum_{k^{'}=1}^{K}} \exp (\rho^{T}_{k^{'}} E_i) }
\end{equation}
where $p \left ( k|E_i \right )$ is the probability that $i$-th pixel being assigned to class $k$, $\rho_k$ is the prototype for $k$-th class and $K$ is the number of classes for semantic segmentation.

\paragraph{Pixel Classification Prototype.} Inspired by traditional semantic segmentation works\cite{long2015fully, chen2017deeplab}, prevalent RIS methods adopt the pixel classification framework to generate the segmentation mask. Specifically, these methods use a linear classifier, \textit{i.e.}, a fully connected layer, to assign each pixel embedding $E_i \in \mathbb{R}^{D}$ a semantic label:

\begin{equation}
\label{equ:pixel classification}
p \left ( k|E_i \right ) = \frac{\exp (w^{T}_k E_i)}{ {\textstyle \sum_{k^{'}=1}^{K}} \exp (w^{T}_{k^{'}} E_i) }
\end{equation}
where $w_k$ is the learned weights for the $k$-th class in the linear classifier. According to Eq~\ref{equ:prototype_view} and Eq~\ref{equ:pixel classification}, the pixel classification framework views the weights of the linear classifiers as the prototype.
In the RIS task, the number of class $K$ is 2 for only one foreground object needs to be segmented,
 which means this mechanism seeks to learn a global prototype for open-world text-image pairs and generates segmentation results by grouping pixels with high responses. We argue that this approach is insufficient for the RIS task since each input text-image pair belongs to a unique class, and using one global prototype to deal with infinite text-image instances is insufficient. 

\paragraph{Learnable Query-based Prototype.} Different from the pixel classification framework, the recent query-based framework generates segments by applying the online updateable prototypes to group pixel embeddings. In contrast to the pixel classification framework that uses a global prototype to fit various text-image pairs, the learnable query-based framework initializes $K$ (\eg, 100) learnable prototypes in the embedding space, and updates them with the multi-modal features for more accurate segmentation. This online updating process of $\rho_k$ can be formulated as:
\begin{equation}
\label{equ:query classification}
\rho_k = q_k + g(q_k,V, L)
\end{equation}
where $q_k$ is $k$-th learned queries, $V$ is the visual input and $L$ is the linguistic input. $g\left ( \cdot  \right ) $ represents cross-modal interactions that generate residuals to update prototypes towards a more proper position according to the input text-image pair. The semantic label of each pixel can be obtained by applying Eq.~\ref{equ:prototype_view}. Compared to a pixel classification-based framework that learns a fixed prototype, the prototypes in a query-based framework can be online updated by multi-modal features. But the prototype $\rho_k$ is still constrained by the fixed learned query $q_k$ and fails to represent the various text-image instances, especially when the language expression is complicated.

{\begin{figure*}[!htb]
\centering
\includegraphics[scale=0.4]{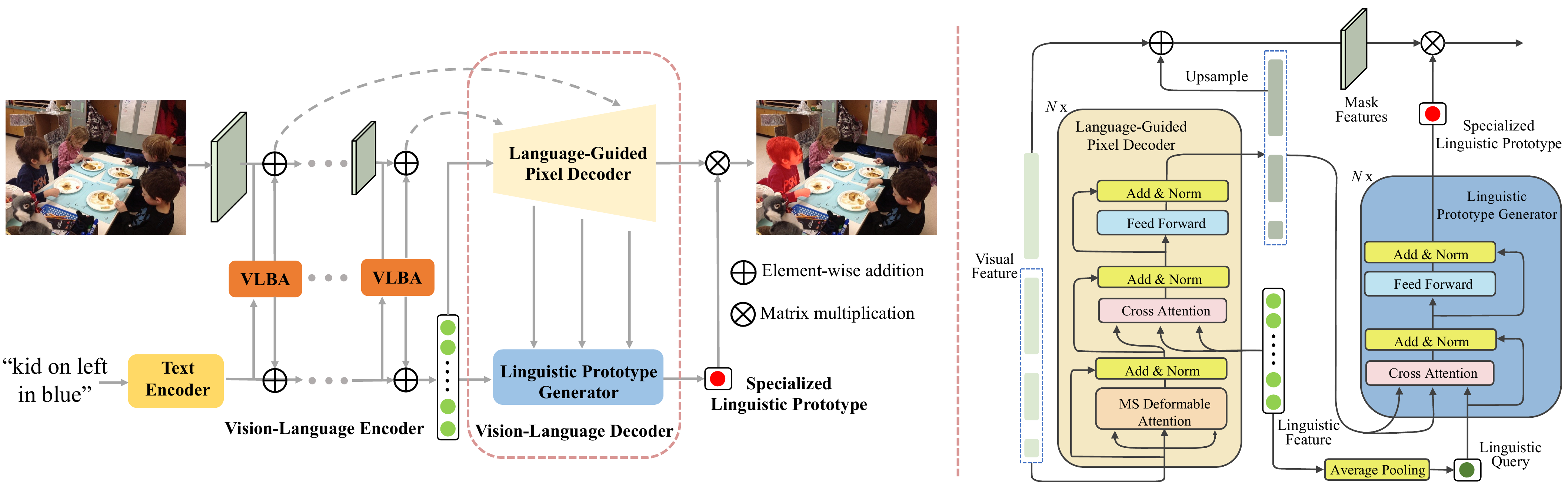}
\caption{An overview of the LGFormer framework is shown on the left side of the figure. It consists of a vision-language encoder with our proposed VLBA module for cross-modal interactions, a vision-language decoder with a language-guided pixel decoder for integrating multi-scale visual features and linguistic features, and a linguistic prototype generator for generating the visual-attended linguistic prototypes. The more detailed design of the vision-language decoder is shown on the right side of the figure.}
\label{Fig.pipeline} %用于文内引用的标签
\vspace{-0.3cm}
\end{figure*}}

\paragraph{Linguistic Prototype.} To deal with various text-image inputs, an intuitive idea is to increase the number of learnable prototypes discussed above. However, too many prototypes may confuse the model associating them with input instances, and make it difficult for the network to converge, as demonstrated in\cite{carion2020end}. Here raises a natural question: \textit{"Is there an efficient way to handle open-world text-image pairs with fewer queries?"}

This work answers this question by incorporating the linguistic features into the query vector to generate a linguistic prototype for each text-image instance. 
This specialized linguistic prototype is initialized by linguistic prior and then further updated with multi-modal information as follow:
\begin{equation}
\label{equ:linguisitic_proto}
\rho = f(L) + g(f(L),V,L)
\end{equation}
where $L$ is the input text and $f\left ( \cdot  \right ) $ represent a text encoder. As such, the specialized linguistic prototype can capture instance-level information of the input, thus achieving more accurate segmentation. The proposed mechanism can handle the arbitrary number of classes with only one non-parametric query vector conditioned on linguistic information, which can fit the RIS task with open class naturally.

\subsection{Network Architecture}

As illustrated in Fig.~\ref{Fig.pipeline}, the proposed LGFormer takes a text-image pair as input and outputs a segmentation mask for the region of interest. Firstly, the input image and text are transformed into multi-modal features, which leverage the proposed VLBA for early cross-modal fusion. Secondly, the multi-scale visual features and linguistic features are fed into the pixel decoder to generate fine-grained semantic representations. Then, the multi-scale visual features are sent to the linguistic prototype generator to update the linguistic query continuously. As such, the instance-level information of the input text-image pair can be captured by the linguistic query and it will act as the specialized prototype for pixels grouping. Finally, the mask is obtained by the dot product between the linguistic prototype and the mask features refined by the pixel decoder. 

\subsubsection{Vision-Language Encoder}

\paragraph{Vision Backbone.} For the input image, we adopt a deep hierarchical vision model (\eg, Swin Transformer~\cite{liu2021swin}) to obtain multi-scale visual features maps with a resolution of 1/4, 1/8, 1/16, and 1/32 of the original spatial size. These feature maps contain rich details and semantics, which will facilitate better cross-modal alignment.

\paragraph{Language Backbone.} For the input language expression with $T$ words, we first use a tokenizer (\eg, BERT Tokenizer~\cite{devlin2018bert}) to transform them into a set of word embeddings, and then adopt a deep language model (\eg, BERT~\cite{devlin2018bert}) to generate linguistic features which will be used for every subsequent process in our model.

{\begin{figure}[!htb]
\centering\includegraphics[scale=0.35]{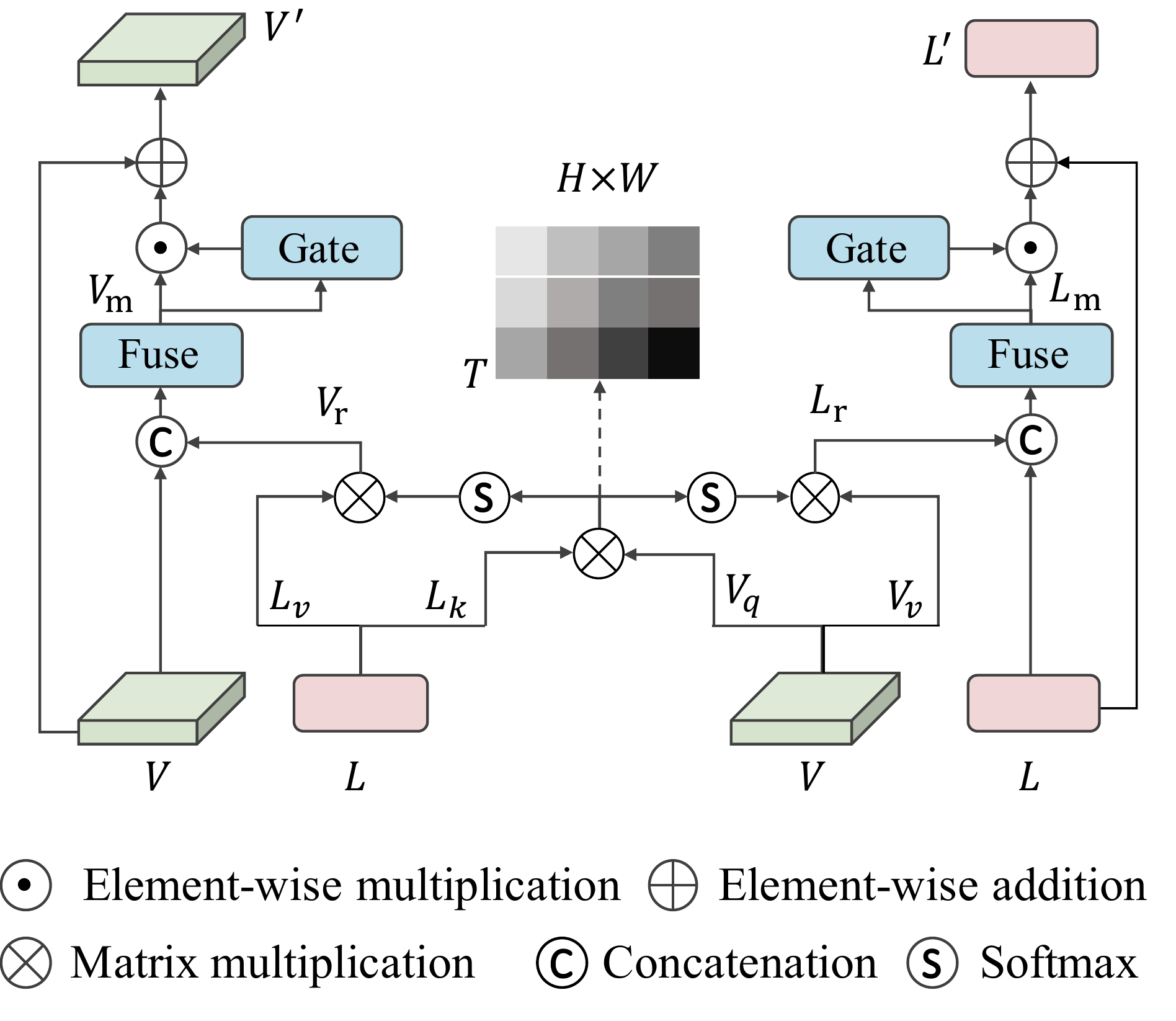}
\caption{The detailed process of our proposed visual-language bidirectional attention (VLBA) module with a symmetric structure. The module takes linguistic features $L$ and visual features $V$ as input and updates them mutually.}
\label{Fig.fusion} %用于文内引用的标签
\vspace{-0.5cm}
\end{figure}}

\paragraph{Vision-Language Bidirectional Attention.} We propose a vision-language bidirectional attention module (\textbf{VLBA}) for early cross-modal alignment based on PWAM~\cite{yang2022lavt}. In contrast to PWAM which views linguistic information as supplementary, the proposed VLBA regards linguistic features as equally important as visual features.
% , and interweaves these two modality features deeply for better alignment. 
As shown in Fig.~\ref{Fig.fusion}, the proposed VLBA takes the visual features $V \in \mathbb{R}^{H \times W \times C_v}$ from the intermediate layer of visual backbone and the linguistic features $L \in \mathbb{R}^{T \times C_l}$ from the last layer of linguistic backbone as inputs, where $H$, $W$, $C_v$, $T$ and $C_l$ are the height, width, channels of the visual feature maps, the word length and the channels of linguistic feature maps. And then update $V$  and $L$ as follow:
% First, bidirectional cross-attention is employed to aggregate relevant cross-modal information according to the similarities of visual and linguistic features. This process is formulated as follows
\begin{equation} A = \frac{V_qL^T_k}{\sqrt{C_v}}
\end{equation}
\begin{equation} V_r =  w_{1} \left (softmax \left ( A  \right )L_v \right )  \end{equation}
\begin{equation} L_r =  w_{2} \left (softmax \left ( A^T  \right )V_v \right ) \end{equation}
where $V_q$ and $V_v$ are the query embeddings and value embeddings of $V$. $L_k$ and $L_v$ are the key embeddings and value embeddings of $L$. $w_1$ and $w_2$ represent linear projections. $V_r$ and $L_r$ are the responses for visual features and linguistic features, respectively. 

% Second, $V_r$ and $L_r$ are fused into original features in the form of residuals to avoid interference caused by irrelevant compensation information:
\begin{equation} V_m = fuse(cat \left( V,V_r \right)) \end{equation}
\begin{equation} L_m = fuse(cat \left( L,L_r \right))  \end{equation}
\begin{equation} V^{'} = V + V_m \odot gate\left ( V_m \right) \end{equation}
\begin{equation} L^{'} = L + L_m \odot gate\left ( L_m \right)  \end{equation}
where $\odot $ denotes element-wise multiplication. Both the $fuse\left ( \cdot  \right ) $ and $gate\left ( \cdot  \right ) $ are implemented as a two-layer MLP.

\subsubsection{Vision-Language Decoder}
We design a cross-modal decoder to perform linguistic prototype-based decoding, which consists of four components: language-guided pixel decoder, vision-guided linguistic prototype generator, and prediction head.
% In contrast to the parametric decoding approach (\textit{i.e.}, pixel classification-based and learnable query-based) adopted in existing methods, the proposed linguistic prototype-based decoding mechanism is more flexible and effective for RIS tasks. 
% Specifically, the specialized linguistic prototype is able to capture instance-level information of the input text-image pair, so it can act as the prototype which will be used to group pixels with high responses into segmentation mask directly without any matching and integration operations. 
% Together with the proposed cross-modal interaction modules (\eg, VLBA) for feature alignment, the proposed linguistic prototype-based decoding mechanism can generate more consistent segmentation results.

% {\begin{figure}[!htb]
% \centering
% \includegraphics[scale=0.45]{figures/decoder.pdf}
% \caption{The design of the vision-language decoder.}
% \label{Fig.decoder} %用于文内引用的标签
% \end{figure}}
\paragraph{Language-Guided Pixel Decoder.} The pixel decoder is used to gradually recover the spatial resolution of visual features. To make per-pixel feature representations more discriminative for mask generation, we also incorporate the linguistic features in this process which will enhance instance features of interest and weaken irrelevant background features. As such, the uni-modal pixel decoder is converted into a cross-modal pixel decoder, performing spatial resolution recovery and multi-scale cross-modal fusion simultaneously.

As shown in Fig.~\ref{Fig.pipeline} right, our pixel decoder consists of $N$ transformer layers and a upsample layer for generating mask features. In each transformer layer, multi-scale deformable attention~\cite{zhu2020deformable}, cross attention~\cite{vaswani2017attention}, and feed-forward network are employed sequentially for feature refinement. 
Specifically, the transformer layers take the multi-scale visual feature maps with resolutions 1/8, 1/16, and 1/32 of original spatial size and linguistic features from the vision-language encoder as inputs, and outputs the refined multi-scale visual feature maps which have the same spatial resolutions as the inputs. 
After this multi-scale cross-modal features fusion, the bilinear interpolation is applied to the feature maps with a resolution 1/8 to obtain the mask features for mask generation.
% with a resolution 1/4 of the original spatial size which will be used to generate the segmentation mask.

\paragraph{Linguistic Prototype Generator.} This generator is used to transform the sentence-level linguistic query to the specialized linguistic prototype for grouping the pixel embedding, which is implemented as a $N$-layer standard transformer decoder~\cite{vaswani2017attention}. Through cross-modal interactions with visual features, the linguistic prototype is associated with the unique referred object, making it easier to handle arbitrary text-image pairs. 

To generate the specialized prototype for the input text-image instance, we first obtain the sentence-level embedding of linguistic features by average pooling across word dimensions. 
Then, this sentence-level embedding together with the multi-scale visual features with resolution 1/32, 1/16, and 1/8 of the original image from the pixel decoder are fed into the linguistic prototype generator. We follow ~\cite{cheng2022masked} to update the linguistic prototype with multi-scale visual features in a sequential manner.
% After these cross-modal interactions, the instance-level information of the text-image instance is captured by the linguistic prototype, and the consequent specialized prototype will be used for segment generation in the subsequent prediction process.

\paragraph{Prediction Head.} A lightweight head is built on top of the linguistic prototype generator to further transform the prototype. This head is implemented as a 3-layer MLP with ReLU activation except for the last layer. Given the specialized linguistic prototype and the mask features with a resolution 1/4 of the original spatial size from the pixel decoder, the segmentation mask is generated by their dot product.

\begin{table*}[htbp]
    \centering
    \setlength{\tabcolsep}{2.3mm}{
    \scalebox{1.}{
        \begin{tabular}{c|c|c|ccc|ccc|cc}
        \Xhline{1pt}
        \multirow{2}{1cm}{\centering Method}&
        \multirow{2}{1.3cm}{\centering Backbone} &
        \multicolumn{1}{c|}{ReferIt} &
        \multicolumn{3}{c|}{RefCOCO}&\multicolumn{3}{c|}{RefCOCO+}&\multicolumn{2}{c}{RefCOCOg}\\
        \cline{3-11} & &test&val&testA&testB&val&testA&testB&val(U)&test(U)\\   
        \hline 
        BRINet~\cite{hu2020bi}          & ResNet-101& 63.46& 60.98 &62.99 &59.21 &48.17 &52.32 &42.11 &- &- \\
        CMPC~\cite{huang2020referring}  & ResNet-101& 65.53& 61.36 &64.53 &59.64& 49.56& 53.44& 43.23& - &- \\
        LSCM~\cite{hui2020linguistic}   & ResNet-101& 66.57& 61.47 &64.99 &59.55 &49.34 &53.12 &43.50 &- &- \\
        CMPC+~\cite{liu2021cross}       & ResNet-101& -& 62.47 &65.08 &60.82 &50.25 &54.04 &43.47 &- &- \\ 
        MCN~\cite{luo2020multi}         & Darknet-53& -& 62.44 &64.20 &59.71 &50.62 &54.99 &44.69 &49.22 &49.40 \\
        EFN~\cite{feng2021encoder}      & ResNet-101& 66.70& 62.76 &65.69 &59.67 &51.50 &55.24 &43.01 &- &- \\
        BUSNet~\cite{yang2021bottom}    & ResNet-101& -& 63.27 &66.41 &61.39 &51.76 &56.87 &44.13 &- &- \\
        CGAN~\cite{luo2020cascade}      & ResNet-101& -& 64.86 &68.04 &62.07 &51.03 &55.51 &44.06 &51.01 &51.69 \\
        LTS~\cite{jing2021locate}       & DarkNet-53& -& 65.43 &67.76 &63.08 &54.21 &58.32 &48.02 &54.40 &54.25 \\
        VLT~\cite{ding2021vision}       & DarkNet-53& -& 65.65 &68.29 &62.73 &55.50 &59.20 &49.36 &52.99 &56.65 \\
        ReSTR~\cite{kim2022restr}       & ViT-B-16& 70.18& 67.22 &69.30 &64.45 &55.78 &60.44 &48.27 &54.48 &- \\
        CRIS~\cite{wang2022cris}        & ResNet-101& -& 70.47 &73.18 &66.10 &62.27 &68.08 &53.68 &59.87 &60.36 \\
        LAVT~\cite{yang2022lavt}        & Swin-B & -& 72.73& 75.82& 68.79& 62.14& 68.38& 55.10& 61.24& 62.09\\
        CoupAlign~\cite{zhang2022coupalign}& Swin-B & 73.28& \textbf{74.70}& 77.76& 70.58& 62.92& 68.34& 56.69& 62.84& 62.22 \\
        \hline
        \textbf{LGFormer (ours)}        & Swin-B & \textbf{75.80}& 74.69 & \textbf{77.81} & \textbf{70.66} & \textbf{65.69} & \textbf{71.53} & \textbf{57.89} & \textbf{63.72} & \textbf{65.18} \\
        \Xhline{1pt}
        \end{tabular}
        }
    }
    \caption{Comparisons with state-of-the-art methods.}
    \vspace{-0.4cm}
    \label{tab:sota}
\end{table*}

\subsection{Vision-Language Contrastive Learning} In contrast to existing pixel classification-based methods which perform feature alignment implicitly with classification loss, our linguistic prototype-based framework can easily introduce contrastive learning for explicit feature alignment, thus achieving better segmentation results.

Given the specialized linguistic prototype $L_o \in \mathbb{R}^{C_o}$ and mask features $V_o \in \mathbb{R}^{N_o \times C_o}$ from the decoder, the pixel-text contrastive loss is formulated as:

\begin{equation}
\label{equ:lpt_i}
\mathcal{L}^{i}
= \\
\begin{cases}
 - \log_{}{\sigma \left ( \left (L_o \cdot V^i_o \right ) \right ) } \qquad \quad \ \ \  if \ y^i=1 \\
- \log_{}{\left (1 - \sigma \left ( \left (  L_o \cdot V^i_o\right ) \right ) \right) } \quad otherwise
\end{cases}
\vspace{-0.1cm}
\end{equation}

\begin{equation}
\label{equ:lpt}
\mathcal{L}
= \frac{1}{N_o} \sum_{i=1}^{N_o} \mathcal{L}^{i}\left ( L_o,V^{i}_o \right )
\end{equation}
where $\sigma \left ( \cdot  \right ) $ is the sigmoid function. $N_o$ and $C_o$ are the spatial size and channels of the visual feature maps $V_o$. $y^i \in \left \{0, 1 \right \}$ and $V^{i}_o$ are the ground truth label and the per-pixel embeddings at the $i$-th position of $V_o$.

The contrastive loss pulls pixel embeddings in the region of interest to the linguistic prototype and pushes that of the background away from it.
% , achieving more fine-grained cross-modal information propagation and feature alignment.

\section{Experiments}

\subsection{Datasets and Metrics}
\label{sec:datasets and metrics}

\paragraph{Datasets.} We evaluate the proposed method on four commonly used datasets including ReferIt~\cite{kazemzadeh2014referitgame}, RefCOCO~\cite{yu2016modeling}, RefCOCO+~\cite{yu2016modeling} and RefCOCOg~\cite{mao2016generation}. ReferIt~\cite{kazemzadeh2014referitgame} contains 19,894 images with 130,525 language expressions that refer to 96,654 object regions. Language expressions in ReferIt are relatively shorter than the other datasets. RefCOCO~\cite{yu2016modeling} contains 19,994 images and 142,209 language expressions that refer to 50,000 segmented object regions. The average length of language expressions in this dataset is 3.5 words. RefCOCO+~\cite{yu2016modeling} contains 26,711 images and 104,560 language expressions that refer to 54,822 segmented object regions. Especially, expressions in RefCOCO+ do not contain words describing the absolute locations of objects, which makes it more challenging than RefCOCO. RefCOCOg~\cite{mao2016generation} consists of 104,560 expressions involving 54,822 objects in 26,711 images. In contrast to the above three datasets, RefCOCOg has a longer average sentence length of 8.4 words, containing more words about the appearance and location of the referent.

\paragraph{Metrics.}We adopt three metrics to evaluate the proposed methods including overall intersection-over-union (oIoU), mean intersection-over-union (mIoU), and $Precision$@$X$. The oIoU is defined as the ratio of the total intersection area over the total union area between the predictions and ground truth of all test data, which reflects the overall segmentation accuracy. The mIoU calculates the average IoU of all test data, which reflects the generalization ability of the model. $Precision$@$X$ measures the percentage of test data that exceeds the predefined IoU threshold $X$, and in this paper $ X\in \left \{ 0.5, 0.7, 0.9 \right \} $.

{\begin{figure*}[!htb]
\centering
\includegraphics[scale=0.38]{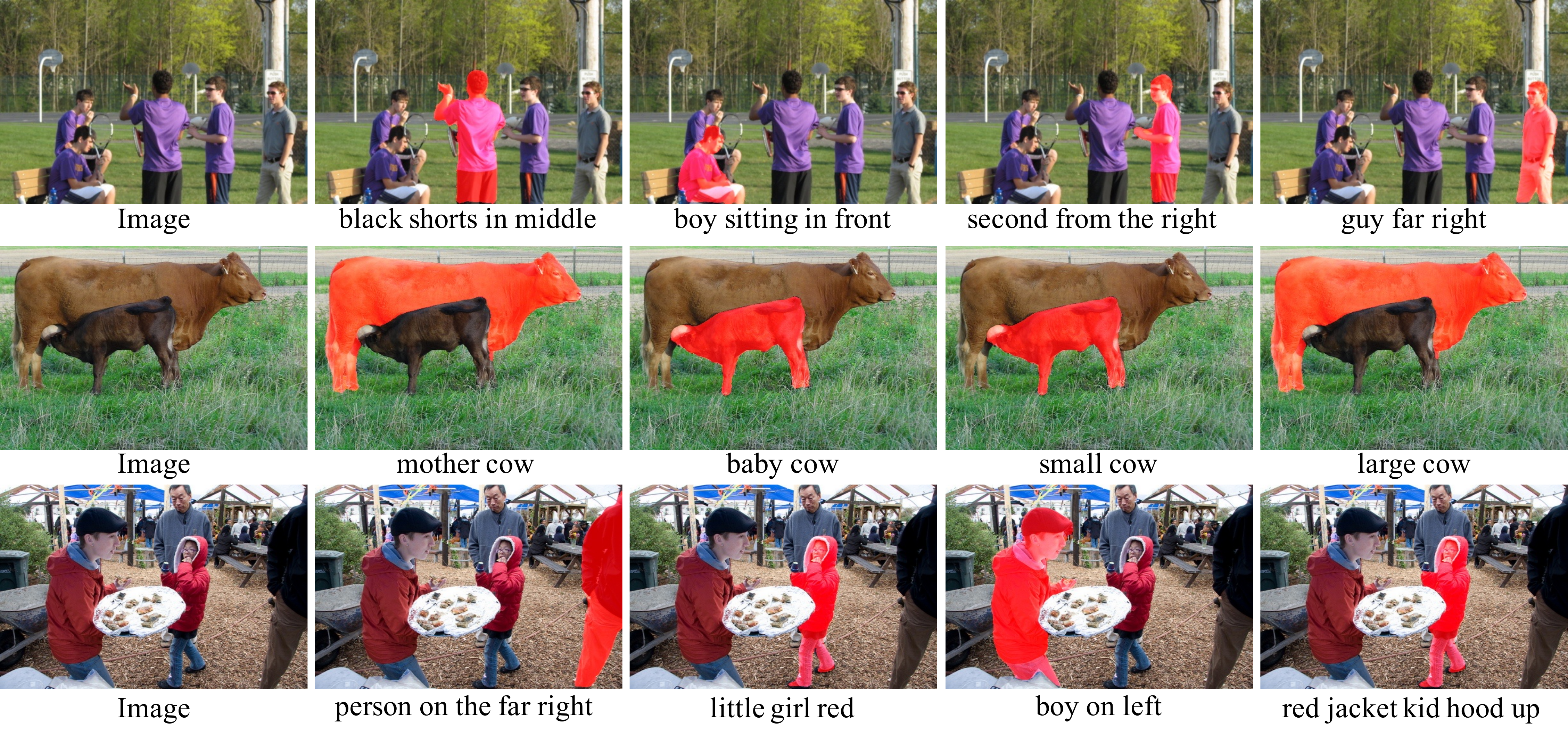}
\caption{Visualization examples of LGFormer on the RefCOCO val set.}
\label{Fig.vis_result} %用于文内引用的标签
\end{figure*}} 

\subsection{Implementation Details.}
\label{implementation details}

We utilize PyTorch to implement our method. To make a fair comparison with the state-of-the-art methods, official pre-trained Swin-Transformer-Base~\cite{liu2021swin} and BERT-Base-Uncased~\cite{devlin2018bert} models are adopted as the visual encoder and linguistic encoder respectively, and the other parameters in our model are initialized randomly. For network optimization, we adopt the AdamW optimizer with a weight decay of 0.01 and an initial learning rate of 5e-5 with the polynomial scheduler. The decoder layer number $N$ is set to 6 by default. All input images are resized to $480 \times 480$ and no data augmentation is applied. We train the model for 40 epochs, and the batch size is set to 32.

During inference, the predicted results with the spatial stride 4 are interpolated into the original spatial size and then binarized with the threshold of 0.5 as the final segmentation mask without any post-process. 

\subsection{Comparisons with State-of-the-art}

In Tab.~\ref{tab:sota}, we compare our method with existing state-of-the-art methods on four widely used datasets detailed in Sec.~\ref{sec:datasets and metrics}. It can be seen that the our LGformer outperforms other methods on almost all datasets with large margins. 

On ReferIt, LGFormer achieves the oIoU of 75.80$\%$ without bells and whistles, which is 2.52\% higher than the previous state-of-the-art method CoupAlign~\cite{zhang2022coupalign}. Similarly, our method also achieves the best results on RefCOCOg and RefCOCO+. In particular, on the more complex and difficult RefCOCO+, LGFormer still achieves a oIoU gain of more than 1.2\%, especially on the testA split, which is 3.19\%. On RefCOCOg, another dataset with a longer average length of language expressions, our method also achieves notable oIoU improvements over other methods by 2.96\% at most. These results show that the proposed linguistic query-based mechanism is able to capture instance-level information through cross-modal interactions, making it more flexible and effective in difficult cases. We visualize some segmentation results on the RefCOCO val subset in Fig.~\ref{Fig.vis_result}, where each image is associated with different language expressions. These visualization results further demonstrate the effectiveness of the proposed method in handling variations of images and language.

\subsection{Ablation Study}
In this section, we investigate the effect of the core components in our model. The more challenging val split with complex scenarios of RefCOCO+~\cite{yu2016modeling} is adopted. Experimental settings are the same as Sec.~\ref{implementation details}.

% \begin{table}\centering
%     \setlength{\tabcolsep}{2.5mm}{
%     \scalebox{1.}{
%     \begin{tabular}{c|c|cc}
%     \Xhline{1pt}  query type& query number & oIoU & mIoU\\
%     \hline
%     \multirow{2}{1cm}{\centering Learnable query}& 1 & 63.15 &66.38\\
%     \cline{2-4} & 10 & 64.26 & 68.14\\
%     \hline
%     Linguistic query  & 1 &\textbf{65.69}&\textbf{69.44}\\
%     \Xhline{1pt}
%     \end{tabular}
%     }
%     }
%     \caption{Ablation studies of query type.}
%     \label{tab:query type}
% \end{table}

\paragraph{Linguistic Query vs. Learnable Query.} We study the effects of different query types on performance, including the proposed linguistic query and learnable queries, as shown in Tab.~\ref{tab:query type}, where we use a 3-layer MLP to integrate the multiple masks when the learnable query number is 10. From the first two rows of Tab.~\ref{tab:query type}, we can see that a larger query number is helpful to boost the performance for capturing more details of the input text-image pair. From the last row of Tab.~\ref{tab:query type}, the proposed linguistic query-based mechanism surpasses the learnable queries-based counterparts by large margins, which shows that our approach can capture instance-level information of the input text-image pair for accurate mask generation.

\begin{table}\centering
    \setlength{\tabcolsep}{1mm}{
    \scalebox{1.}{
    \begin{tabular}{c|c|ccc|cc}
    \Xhline{1pt}  Query Type& Num & P@0.5&P@0.7&P@0.9 & oIoU & mIoU\\
    \hline
    \multirow{2}{1.4cm}{Learnable}& 1 & 75.59& 67.48& 29.37& 63.15 &66.38\\
    \cline{2-7} & 10 & 77.37& 69.64& 31.96& 64.26 & 68.14\\
    \hline
    Linguistic & 1 & \textbf{78.96}& \textbf{71.30}& \textbf{33.35} &\textbf{65.69}&\textbf{69.44}\\
    \Xhline{1pt}
    \end{tabular}
    }
    }
    \caption{Ablation studies of query type.}
    % \vspace{-0.8cm}
    \label{tab:query type}
\end{table}

In order to further analyze the ability of the linguistic query for the complicated expressions, we divide the RefCOCO+(val) into four parts according to the length of expressions $l_{text}$ (the long expressions are usually more complicated than the short expressions) and evaluate the oIoU respectively. Specifically, we divide $1\le l_{text}<3$, $3\le l_{text}<8$, $8\le l_{text}<13$ and $l_{text}\ge13$ as \textit{simple} set, \textit{moderate} set, \textit{complicated} set, and \textit{extremely complicated} set. Tab.~\ref{tab:linguistic_analyze} shows that the linguistic query-based method outperforms the learnable query-based method by 0.6$\%$, 1.3$\%$, 2.5$\%$ and \textbf{7.7$\%$} under the oIoU metric on the \textit{simple} set, \textit{moderate} set, \textit{complicated} set, and \textit{extremely complicated} set respectively. The superiority of the linguistic query-based method becomes more obvious as the linguistic complexity increases. We visualize some segmentation results of the cases with complicated language expressions in Fig.~\ref{Fig.ling_vis_result}, the learnable query-based method is restricted by the query number and thus fails to fit open-world text-image pairs with diverse scenarios. In contrast, our linguistic prototype-based mechanism is able to capture the instance-level information of the input text-image instance, thus generating a more consistent mask even in complicated scenarios.

{\begin{figure}[!htb]
\centering
\includegraphics[scale=0.75]{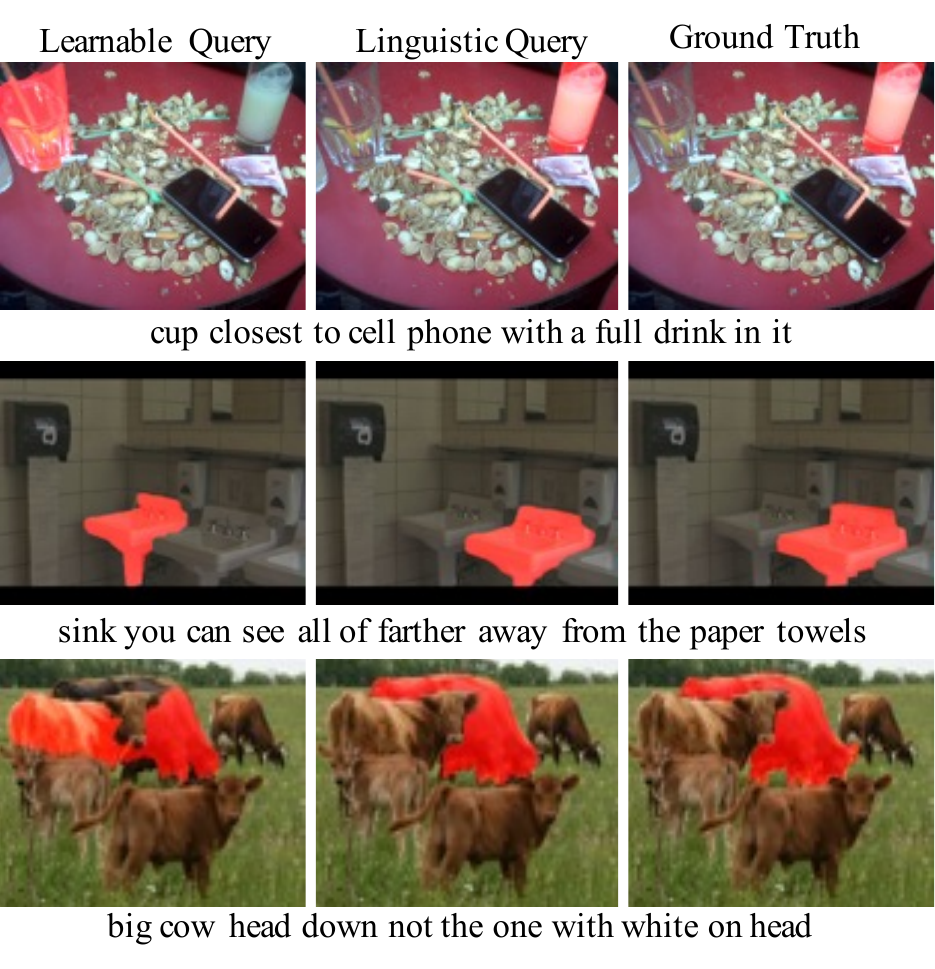}
\caption{Visualization of the cases with complicated expressions.}
    \vspace{-0.8cm}
\label{Fig.ling_vis_result} %用于文内引用的标签
\end{figure}} 

\begin{table}\centering
    \setlength{\tabcolsep}{1.5mm}{
    \scalebox{1.}{
        \begin{tabular}{c|cccc}
        \Xhline{1pt}  \diagbox{Method}{Length}& [1,3) & [3,8)& [8,13)& [13,+$\infty$) \\
        \hline
        Learnable& 75.20 & 64.11& 51.64& 43.13\\
        \hline
        Linguistic & \textbf{75.86} & \textbf{65.33}& \textbf{54.61}& \textbf{50.81}\\
        \Xhline{1pt}
        \end{tabular}
        }
    }
    \caption{Ablation studies of complicated expressions.}
    \vspace{-0.5cm}
    \label{tab:linguistic_analyze}
\end{table}
\paragraph{Cross-modal Fusion Modules.} Tab.~\ref{tab:components} shows the effect of different cross-modal fusion modules on the performance, including vision-language bidirectional attention (VLBA) at the encoder, language-to-vision attention (L2VA) and vision-to-language attention (V2LA) at the decoder. The baseline (i.e., first row) is built by removing L2VA and V2LA and replacing the VLBA with PWAM proposed in~\cite{yang2022lavt}. It can be seen that our baseline has already achieved strong performance, exceeding other state-of-the-art methods with obvious margins. This is because our linguistic query-based mechanism is able to capture instance-level information from visual and linguistic inputs, thus promoting more consistent and accurate segmentation results. 

From the second row of Tab.~\ref{tab:components}, the VLBA module brings $0.9\%$ oIoU improvements. We can also see that cross-modal fusion modules L2VA and V2LA at the decoder further boost the oIoU performance by $0.41\%$ and $0.47\%$ respectively. These results demonstrate that cross-modal interactions in both the encoder and decoder are essential for achieving better alignment.

\begin{table}\centering
    \setlength{\tabcolsep}{3mm}{
    \scalebox{1.}{
    \begin{tabular}{ccc|cc}
    \Xhline{1pt}  VLBA& L2VA & V2LA & oIoU & mIoU\\
    \hline
                &           &           &63.91&67.47\\
    \checkmark  &           &           &64.81&68.68\\
    \checkmark  &\checkmark &           &65.22&68.75\\
    \checkmark  &\checkmark &\checkmark &\textbf{65.69}&\textbf{69.44}\\
    \Xhline{1pt}
    \end{tabular}
    }
    }
    \caption{Ablation studies of cross-modal fusion modules.}
    \label{tab:components}
\end{table}

\paragraph{Number of Decoder Layers.} To investigate the effect of the number of decoder layers, we vary it to $\{ 3, 6\}$. Note that the layer numbers of L2VA and V2LA are kept the same in this study. As shown in Tab.~\ref{tab:decoder layer}, the model performance is enhanced consistently with more layers. We choose 6 as the default decoder layer number.

\begin{table}\centering
    \setlength{\tabcolsep}{6mm}{
    \scalebox{1.}{
    \begin{tabular}{c|cc}
    \Xhline{1pt}  Layer Num& oIoU & mIoU\\
    \hline
    3 & 65.31 & 69.02\\
    6 & \textbf{65.69} & \textbf{69.44}\\
    \Xhline{1pt}
    \end{tabular}
    }
    }
    \caption{Ablation studies of decoder layer's number.}
    \label{tab:decoder layer}
\end{table}

\paragraph{Word Embeddings vs. Sentence Embeddings for L2VA.} For the L2VA in the language-guided pixel decoder, the linguistic features can be used in two forms, i.e., original word embeddings and sentence embeddings, where the sentence embeddings are generated by the average pooling across word dimensions. As shown in Tab.~\ref{tab:word_sentence}, using the word embeddings can outperform using sentence embedddings by $0.33\%$ oIoU gain. This is because fine-grained linguistic features can promote better alignment than global features.

\begin{table}\centering
    \setlength{\tabcolsep}{4mm}{
    \scalebox{1.}{
    \begin{tabular}{c|cc}
    \Xhline{1pt}  Query of L2VA& oIoU & mIoU\\
    \hline
    sentence embeddings & 65.36 &68.93\\
    word embeddings & \textbf{65.69} & \textbf{69.44}\\
    \Xhline{1pt}
    \end{tabular}
    }
    }
    \caption{Ablation results on the input of pixel decoder.}
    \vspace{-0.5cm}
    \label{tab:word_sentence}
\end{table}

\subsection{Conclusion}
In this paper, we introduce the linguistic query-guided mask generation for referring image segmentation task for the first time, and propose an end-to-end framework LGFormer. It utilizes the linguistic features as the query to generate a specialized prototype and groups pixels with high responses into the segmentation mask directly. Moreover, we design several cross-modal fusion modules in both encoder and decoder to achieve better alignment, including the VLBA module in the encoder, language-to-vision attention (L2VA) and vision-to-language attention (V2LA) in the decoder. Extensive experiments demonstrate that our LGFormer surpasses previous state-of-the-art methods on multiple benchmark datasets by large margins. 

{\small
\bibliographystyle{ieee_fullname}
\bibliography{egbib}

\begin{thebibliography}{10}\itemsep=-1pt

\bibitem{carion2020end}
Nicolas Carion, Francisco Massa, Gabriel Synnaeve, Nicolas Usunier, Alexander
  Kirillov, and Sergey Zagoruyko.
\newblock End-to-end object detection with transformers.
\newblock In {\em European Conference on Computer Vision (ECCV)}, 2020.

\bibitem{chen2019see}
Ding-Jie Chen, Songhao Jia, Yi-Chen Lo, Hwann-Tzong Chen, and Tyng-Luh Liu.
\newblock See-through-text grouping for referring image segmentation.
\newblock In {\em International Conference on Computer Vision (ICCV)}, 2019.

\bibitem{chen2018language}
Jianbo Chen, Yelong Shen, Jianfeng Gao, Jingjing Liu, and Xiaodong Liu.
\newblock Language-based image editing with recurrent attentive models.
\newblock In {\em Conference on Computer Vision and Pattern Recognition
  (CVPR)}, 2018.

\bibitem{chen2017deeplab}
Liang-Chieh Chen, George Papandreou, Iasonas Kokkinos, Kevin Murphy, and Alan~L
  Yuille.
\newblock Deeplab: Semantic image segmentation with deep convolutional nets,
  atrous convolution, and fully connected crfs.
\newblock {\em IEEE transactions on pattern analysis and machine intelligence
  (PAMI)}, 2017.

\bibitem{cheng2022masked}
Bowen Cheng, Ishan Misra, Alexander~G Schwing, Alexander Kirillov, and Rohit
  Girdhar.
\newblock Masked-attention mask transformer for universal image segmentation.
\newblock In {\em Conference on Computer Vision and Pattern Recognition
  (CVPR)}, 2022.

\bibitem{cheng2021per}
Bowen Cheng, Alex Schwing, and Alexander Kirillov.
\newblock Per-pixel classification is not all you need for semantic
  segmentation.
\newblock {\em Advances in Neural Information Processing Systems (NIPS)}, 2021.

\bibitem{devlin2018bert}
Jacob Devlin, Ming-Wei Chang, Kenton Lee, and Kristina Toutanova.
\newblock Bert: Pre-training of deep bidirectional transformers for language
  understanding.
\newblock {\em arXiv preprint arXiv:1810.04805}, 2018.

\bibitem{ding2021vision}
Henghui Ding, Chang Liu, Suchen Wang, and Xudong Jiang.
\newblock Vision-language transformer and query generation for referring
  segmentation.
\newblock In {\em International Conference on Computer Vision (ICCV)}, 2021.

\bibitem{feng2021encoder}
Guang Feng, Zhiwei Hu, Lihe Zhang, and Huchuan Lu.
\newblock Encoder fusion network with co-attention embedding for referring
  image segmentation.
\newblock In {\em Conference on Computer Vision and Pattern Recognition
  (CVPR)}, 2021.

\bibitem{fu2019dual}
Jun Fu, Jing Liu, Haijie Tian, Yong Li, Yongjun Bao, Zhiwei Fang, and Hanqing
  Lu.
\newblock Dual attention network for scene segmentation.
\newblock In {\em Conference on Computer Vision and Pattern Recognition
  (CVPR)}, 2019.

\bibitem{he2019adaptive}
Junjun He, Zhongying Deng, Lei Zhou, Yali Wang, and Yu Qiao.
\newblock Adaptive pyramid context network for semantic segmentation.
\newblock In {\em Conference on Computer Vision and Pattern Recognition
  (CVPR)}, 2019.

\bibitem{he2017mask}
Kaiming He, Georgia Gkioxari, Piotr Doll{\'a}r, and Ross Girshick.
\newblock Mask r-cnn.
\newblock In {\em International Conference on Computer Vision (ICCV)}, 2017.

\bibitem{hu2016segmentation}
Ronghang Hu, Marcus Rohrbach, and Trevor Darrell.
\newblock Segmentation from natural language expressions.
\newblock In {\em European Conference on Computer Vision (ECCV)}, 2016.

\bibitem{hu2020bi}
Zhiwei Hu, Guang Feng, Jiayu Sun, Lihe Zhang, and Huchuan Lu.
\newblock Bi-directional relationship inferring network for referring image
  segmentation.
\newblock In {\em Conference on Computer Vision and Pattern Recognition
  (CVPR)}, 2020.

\bibitem{huang2020referring}
Shaofei Huang, Tianrui Hui, Si Liu, Guanbin Li, Yunchao Wei, Jizhong Han, Luoqi
  Liu, and Bo Li.
\newblock Referring image segmentation via cross-modal progressive
  comprehension.
\newblock In {\em Conference on Computer Vision and Pattern Recognition
  (CVPR)}, 2020.

\bibitem{hui2020linguistic}
Tianrui Hui, Si Liu, Shaofei Huang, Guanbin Li, Sansi Yu, Faxi Zhang, and
  Jizhong Han.
\newblock Linguistic structure guided context modeling for referring image
  segmentation.
\newblock In {\em European Conference on Computer Vision (ECCV)}, 2020.

\bibitem{jing2021locate}
Ya Jing, Tao Kong, Wei Wang, Liang Wang, Lei Li, and Tieniu Tan.
\newblock Locate then segment: A strong pipeline for referring image
  segmentation.
\newblock In {\em Conference on Computer Vision and Pattern Recognition
  (CVPR)}, 2021.

\bibitem{kamath2021mdetr}
Aishwarya Kamath, Mannat Singh, Yann LeCun, Gabriel Synnaeve, Ishan Misra, and
  Nicolas Carion.
\newblock Mdetr-modulated detection for end-to-end multi-modal understanding.
\newblock In {\em International Conference on Computer Vision (ICCV)}, 2021.

\bibitem{kazemzadeh2014referitgame}
Sahar Kazemzadeh, Vicente Ordonez, Mark Matten, and Tamara Berg.
\newblock Referitgame: Referring to objects in photographs of natural scenes.
\newblock In {\em Conference on Empirical Methods in Natural Language
  Processing (EMNLP)}, 2014.

\bibitem{kim2022restr}
Namyup Kim, Dongwon Kim, Cuiling Lan, Wenjun Zeng, and Suha Kwak.
\newblock Restr: Convolution-free referring image segmentation using
  transformers.
\newblock In {\em Conference on Computer Vision and Pattern Recognition
  (CVPR)}, 2022.

\bibitem{li2018referring}
Ruiyu Li, Kaican Li, Yi-Chun Kuo, Michelle Shu, Xiaojuan Qi, Xiaoyong Shen, and
  Jiaya Jia.
\newblock Referring image segmentation via recurrent refinement networks.
\newblock In {\em Conference on Computer Vision and Pattern Recognition
  (CVPR)}, 2018.

\bibitem{li2021mail}
Zizhang Li, Mengmeng Wang, Jianbiao Mei, and Yong Liu.
\newblock Mail: A unified mask-image-language trimodal network for referring
  image segmentation.
\newblock {\em arXiv preprint arXiv:2111.10747}, 2021.

\bibitem{liu2017recurrent}
Chenxi Liu, Zhe Lin, Xiaohui Shen, Jimei Yang, Xin Lu, and Alan Yuille.
\newblock Recurrent multimodal interaction for referring image segmentation.
\newblock In {\em International Conference on Computer Vision (ICCV)}, 2017.

\bibitem{liu2021cross}
Si Liu, Tianrui Hui, Shaofei Huang, Yunchao Wei, Bo Li, and Guanbin Li.
\newblock Cross-modal progressive comprehension for referring segmentation.
\newblock {\em Pattern Analysis and Machine Intelligence (PAMI)}, 2021.

\bibitem{liu2021swin}
Ze Liu, Yutong Lin, Yue Cao, Han Hu, Yixuan Wei, Zheng Zhang, Stephen Lin, and
  Baining Guo.
\newblock Swin transformer: Hierarchical vision transformer using shifted
  windows.
\newblock In {\em International Conference on Computer Vision (ICCV)}, 2021.

\bibitem{long2015fully}
Jonathan Long, Evan Shelhamer, and Trevor Darrell.
\newblock Fully convolutional networks for semantic segmentation.
\newblock In {\em Conference on Computer Vision and Pattern Recognition
  (CVPR)}, 2015.

\bibitem{luo2020cascade}
Gen Luo, Yiyi Zhou, Rongrong Ji, Xiaoshuai Sun, Jinsong Su, Chia-Wen Lin, and
  Qi Tian.
\newblock Cascade grouped attention network for referring expression
  segmentation.
\newblock In {\em ACM International Conference on Multimedia (ACM MM)}, 2020.

\bibitem{luo2020multi}
Gen Luo, Yiyi Zhou, Xiaoshuai Sun, Liujuan Cao, Chenglin Wu, Cheng Deng, and
  Rongrong Ji.
\newblock Multi-task collaborative network for joint referring expression
  comprehension and segmentation.
\newblock In {\em Conference on Computer Vision and Pattern Recognition
  (CVPR)}, 2020.

\bibitem{mao2016generation}
Junhua Mao, Jonathan Huang, Alexander Toshev, Oana Camburu, Alan~L Yuille, and
  Kevin Murphy.
\newblock Generation and comprehension of unambiguous object descriptions.
\newblock In {\em Conference on Computer Vision and Pattern Recognition
  (CVPR)}, 2016.

\bibitem{margffoy2018dynamic}
Edgar Margffoy-Tuay, Juan~C P{\'e}rez, Emilio Botero, and Pablo Arbel{\'a}ez.
\newblock Dynamic multimodal instance segmentation guided by natural language
  queries.
\newblock In {\em European Conference on Computer Vision (ECCV)}, 2018.

\bibitem{radford2021learning}
Alec Radford, Jong~Wook Kim, Chris Hallacy, Aditya Ramesh, Gabriel Goh,
  Sandhini Agarwal, Girish Sastry, Amanda Askell, Pamela Mishkin, Jack Clark,
  et~al.
\newblock Learning transferable visual models from natural language
  supervision.
\newblock In {\em International Conference on Machine Learning (ICML)}, 2021.

\bibitem{vaswani2017attention}
Ashish Vaswani, Noam Shazeer, Niki Parmar, Jakob Uszkoreit, Llion Jones,
  Aidan~N Gomez, {\L}ukasz Kaiser, and Illia Polosukhin.
\newblock Attention is all you need.
\newblock {\em Advances in Neural Information Processing Systems (NIPS)}, 2017.

\bibitem{wang2021max}
Huiyu Wang, Yukun Zhu, Hartwig Adam, Alan Yuille, and Liang-Chieh Chen.
\newblock Max-deeplab: End-to-end panoptic segmentation with mask transformers.
\newblock In {\em Conference on Computer Vision and Pattern Recognition
  (CVPR))}, 2021.

\bibitem{wang2019reinforced}
Xin Wang, Qiuyuan Huang, Asli Celikyilmaz, Jianfeng Gao, Dinghan Shen,
  Yuan-Fang Wang, William~Yang Wang, and Lei Zhang.
\newblock Reinforced cross-modal matching and self-supervised imitation
  learning for vision-language navigation.
\newblock In {\em Conference on Computer Vision and Pattern Recognition
  (CVPR)}, 2019.

\bibitem{wang2022cris}
Zhaoqing Wang, Yu Lu, Qiang Li, Xunqiang Tao, Yandong Guo, Mingming Gong, and
  Tongliang Liu.
\newblock Cris: Clip-driven referring image segmentation.
\newblock In {\em Conference on Computer Vision and Pattern Recognition
  (CVPR)}, 2022.

\bibitem{wu2022towards}
Jianzong Wu, Xiangtai Li, Xia Li, Henghui Ding, Yunhai Tong, and Dacheng Tao.
\newblock Towards robust referring image segmentation.
\newblock {\em arXiv preprint arXiv:2209.09554}, 2022.

\bibitem{yang2021bottom}
Sibei Yang, Meng Xia, Guanbin Li, Hong-Yu Zhou, and Yizhou Yu.
\newblock Bottom-up shift and reasoning for referring image segmentation.
\newblock In {\em Conference on Computer Vision and Pattern Recognition
  (CVPR)}, 2021.

\bibitem{yang2022lavt}
Zhao Yang, Jiaqi Wang, Yansong Tang, Kai Chen, Hengshuang Zhao, and Philip~HS
  Torr.
\newblock Lavt: Language-aware vision transformer for referring image
  segmentation.
\newblock In {\em Conference on Computer Vision and Pattern Recognition
  (CVPR)}, 2022.

\bibitem{ye2020dual}
Linwei Ye, Zhi Liu, and Yang Wang.
\newblock Dual convolutional lstm network for referring image segmentation.
\newblock {\em IEEE Transactions on Multimedia}, 2020.

\bibitem{ye2019cross}
Linwei Ye, Mrigank Rochan, Zhi Liu, and Yang Wang.
\newblock Cross-modal self-attention network for referring image segmentation.
\newblock In {\em Conference on Computer Vision and Pattern Recognition
  (CVPR)}, 2019.

\bibitem{yu2016modeling}
Licheng Yu, Patrick Poirson, Shan Yang, Alexander~C Berg, and Tamara~L Berg.
\newblock Modeling context in referring expressions.
\newblock In {\em European Conference on Computer Vision (ECCV)}, 2016.

\bibitem{yu2022k}
Qihang Yu, Huiyu Wang, Siyuan Qiao, Maxwell Collins, Yukun Zhu, Hartwig Adam,
  Alan Yuille, and Liang-Chieh Chen.
\newblock k-means mask transformer.
\newblock In {\em European Conference on Computer Vision (ECCV)}, 2022.

\bibitem{zhang2022coupalign}
Zicheng Zhang, Yi Zhu, Jianzhuang Liu, Xiaodan Liang, and Wei Ke.
\newblock Coupalign: Coupling word-pixel with sentence-mask alignments for
  referring image segmentation.
\newblock {\em arXiv preprint arXiv:2212.01769}, 2022.

\bibitem{zhou2022rethinking}
Tianfei Zhou, Wenguan Wang, Ender Konukoglu, and Luc Van~Gool.
\newblock Rethinking semantic segmentation: A prototype view.
\newblock In {\em Conference on Computer Vision and Pattern Recognition
  (CVPR)}, 2022.

\bibitem{zhu2020deformable}
Xizhou Zhu, Weijie Su, Lewei Lu, Bin Li, Xiaogang Wang, and Jifeng Dai.
\newblock Deformable detr: Deformable transformers for end-to-end object
  detection.
\newblock {\em arXiv preprint arXiv:2010.04159}, 2020.

\end{thebibliography}
}

\end{document}